# A straightforward method to assess motion blur for different types of displays

Fuhao Chen, Jun Chen*
State Key Laboratory of Optoelectronic Materials and Technologies, Guangdong Province Key Laboratory of Display Material and Technology, School of Physics and Engineering, Sun Yat-sen University, Guangzhou 510275, China

Feng Huang
Guangzhou Institute of Measurement and Testing Technology, Guangzhou, China

*email: stscjun@mail.sysu.edu.cn

## Abstract

A simulation method based on the liquid crystal response and the human visual system is suitable to characterize motion blur for LCDs but not other display types. We propose a more straightforward and widely applicable method to quantify motion blur based on the width of the moving object. We thus compare various types of displays objectively. A perceptual experiment was conducted to validate the proposed method. We test varying motion velocities for nine commercial displays. We compare the three motion blur evaluation methods (simulation, human perception, and our method) using z-scores. Our comparisons indicate that our method accurately characterizes motion blur for various display types.

## Keywords

# 1 Introduction

Flat panel displays (FPDs) are widely used for various applications such as entertainment, advertisement, and office work. End users expect a high level of display quality not only with respect to luminance, contrast, and color reproducibility, but also with respect to motion display characteristics [1].

Liquid-crystal displays (LCDs) have largely replaced cathode ray tubes (CRTs) in both the desktop monitor market and the TV market [2] as a result of improved resolution, brightness, signal-to-noise ratio, color gamut, and dynamic range, as well as increased size. However, limitations in motion display performance render LCDs less suitable for some applications, particularly in the medical field [3] and psychological experiments.

Motion display properties are influenced by the display type, response time, and driving waveforms used [4]. Motion blur, one of the fundamental properties of video display, occurs when displaying a video of an object moving in a particular direction. The eye tracks the object while the display presents a frame-by-frame representation of the object's changing location. Because the eye continues to move even during a single frame interval, the image is blurred on the retina over the frame period.

Since motion blur is apparent in LCDs, various attempts have been made to measure and evaluate it [5]-[9]. Though motion blur is not as obvious in other displays, it does exist. However, there are no unified objective measurement methods to measure and compare motion blur between many types of monitors. Existing objective metrics to quantify perceived dynamic performance are typically developed independently for each particular display type. Blurred edge width is a common metric but it has limitations since perceived blur is affected by edge enhancement functions, driving frame frequency, and various driving methods such as black band cycling [10]. To cover more of these motion blur measurement characteristics, we propose using the moving block width (MBW) parameter to generate comparable metrics between different types of displays.

In the next section, we introduce motion blur calculations using blurred edge time (BET) and MBW. Section 3 lays out our experimental setup. In section 4, we describe our perception experiment to validate the proposed motion blur measurement method. In section 5, we compare motion blur estimation using BET, a perception questionnaire, and MBW. Sections 6 and 7 provide a discussion and summary of our findings.

# 2 Motion blur quantification

## 2.1 Human perception of motion blur

One of the most important remaining drawbacks in current displays is an artifact called motion blur, which appears when displaying videos depicting motion. When this optical artifact is present, the observer perceives the borders of moving objects as blurred edges [8].

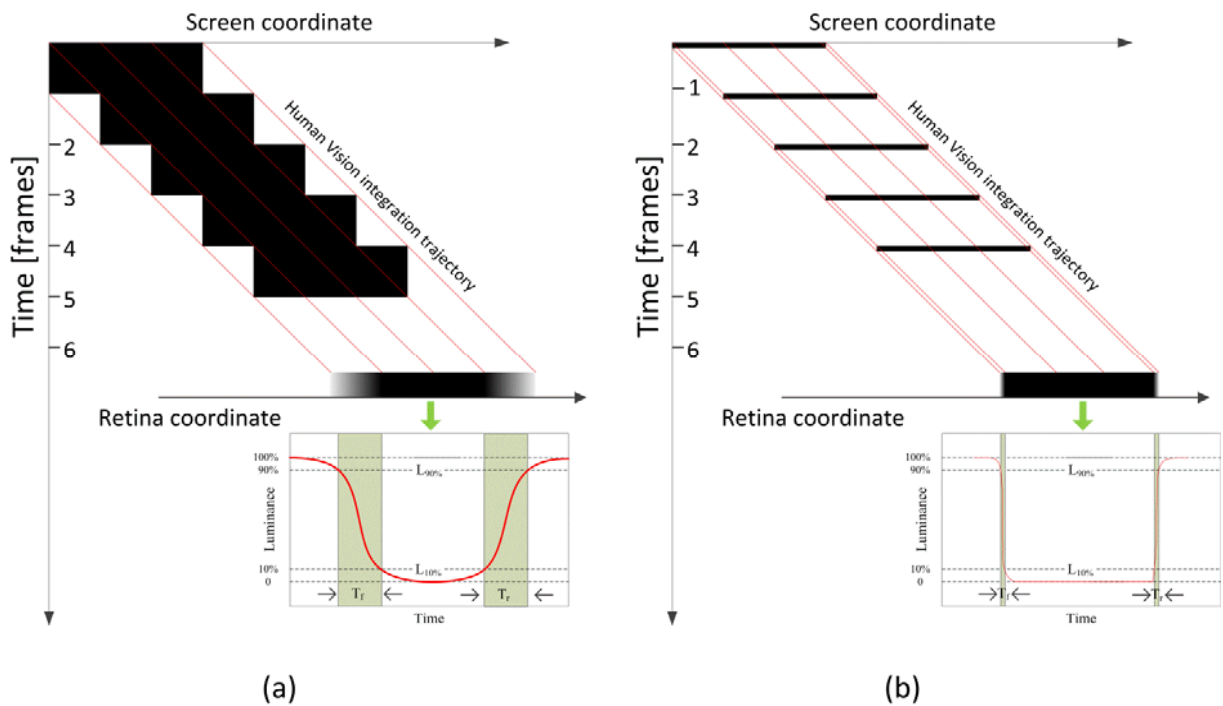

Figure 1- Origin of motion blur. The slanted red lines indicate the trajectory of smooth-pursuit eye tracking. (a) Motion blur perceived from sample-and-hold type displays (e.g., LCDs, PDPs); (b) Motion blur perceived from ideal impulse type displays (e.g., CRTs, LCDs with backlight blinking).

Motion blur is caused by a combination of the light behavior of the display and the human visual system. The temporal response of LCDs is characterized by the sample-and-hold effect whereas CRT displays have a rapid phosphor time decay [11] and plasma display panels (PDPs) have sub-field addressing. Considering a one-dimensional image displayed on a screen, each particular pixel remains illuminated for one frame period in a sample-and-hold type display. Figure 1(a) illustrates a simple example of the eyes tracking one dark moving bar on a bright background. The bar is stationary during each single frame and moves to another position with speed $v$ (pixels/frame) in the next frame. The signal in this example is discrete in both the spatial domain (quantified in pixels) and the temporal domain (quantified in frames) [7]. However, the human eye will track the moving object continuously with a trajectory indicated by the slanted red lines in Figure 1. Due to the lower temporal acuity of the human eye, the light intensity is integrated over one frame period. Therefore, during one frame, the brightness received in the retina involves several pixels. Thus, the observer perceives the borders of moving objects as blurred edges. Figure 1(a) corresponds to an ideal sample-and-hold type display without a response delay. Because of slow liquid crystal response time (LCRT), the motion edge blur is significant. When the image moves from left to right, the pixel on the left side of the edge will be turned OFF from the ON state. The LCRT causes the light to extinguish gradually rather than immediately. For this reason, new driving schemes such as under and over-driving, OCB mode, and storage capacitors have been employed to achieve a faster liquid crystal response. Methods to mitigate the sample-and-hold effect include inserting black data frames and using scanning or blinking backlights. With these improvements to the slow LCRT and the sample-and-hold effect, the LCD is likely to approach the CRT in dynamic display performance (i.e., sample-and-hold type displays can mimic impulse-type displays for motion display). The perceived motion image from an ideal impulse-type display is shown in Figure 1(b).

Comparing the perceived bar width in retina coordinates between Figure 1(a) and (b), we can see the width of the perceived bar correlates with the severity of the blur. For this reason, we use the moving block width as a metric to quantify motion blur. In the following subsections, we introduce LCD motion blur characterization using a camera method and a simulation algorithm. We then propose our motion blur measurement approach, which is applicable to a variety of display types.

## 2.2 Camera measurement method for motion blur

Motion picture response time (MPRT) is used to characterize the motion artifacts of a display. Motion blur is perceived as a result of smooth-pursuit eye tracking of a moving object. The perceived retinal image is obtained by spatial-temporal integration along the motion tracking trajectory. This smooth-pursuit eye tracking can be captured using a smooth-pursuit camera system to simulate moving edge tracking and perform spatial-temporal integration with position shifts [12]. Such a system is set up using a moving camera, a rotating camera, a stationary camera with a rotating mirror to mimic smooth eye tracking, or a high speed camera. The blurred edge width (BEW) is obtained from the measured cross-sectional luminance profile as shown in Figure 2, where BEW is the width of $\overline{ab}+\overline{cd}$.

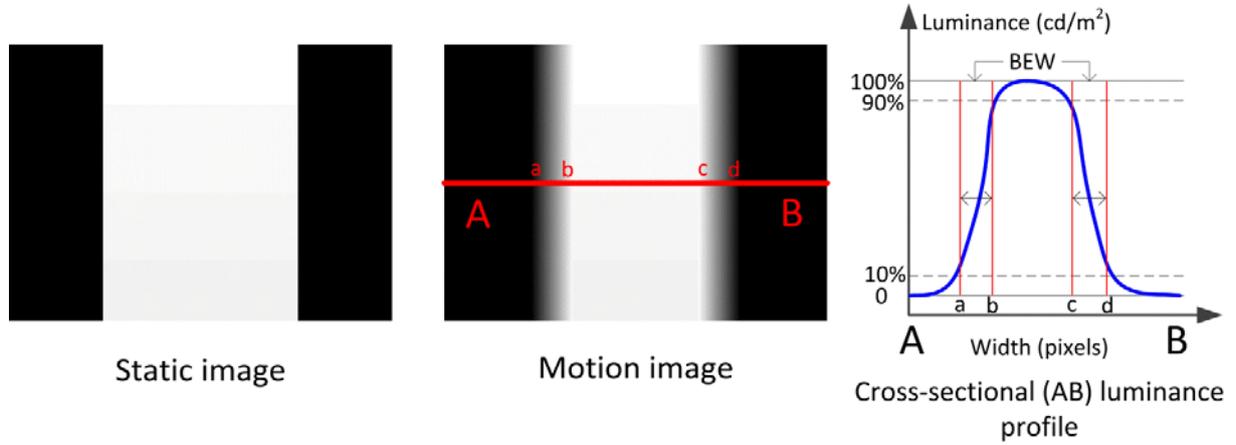

Figure 2- The definition of BEW for a vertical scrolling bar.

The following parameters describe motion blur in terms of the captured luminance profile.

BEW=blurred edge width (in pixels)

$$\text{N-BEW(in frame)} = \frac{\text{BEW}}{\text{moving speed (in pixels/frame)}} \quad (1)$$

N-BET(in sec)=N-BEW×T (in sec/frame)

MPRT(in sec)=average of N-BET

## 2.3 Simulation algorithm for motion blur

As shown in Figure 1(a), the perceived moving picture in retina coordinates is determined by the pixel luminance sum along the motion trajectory during one frame period. When the object moves from left to right, the blur edge is calculated using pixel luminance step response curves combined with eye tracking and temporal integration [13]. The perceived picture can be expressed as shown in Eq. (2):

$$V(x_R) = \frac{1}{T} \sum_{m=0}^{v-1} \int_{mT/v}^{(m+1)T/v} Y(x_S + m, t) dt \quad (2)$$

where $V(x_R)$ is the perceived luminance on the retina position $x_R$ after integrating over one frame $T$. $Y(x_S + m, t)$ is the brightness from pixel $x_S + m$ on the screen at instant $t$ during one frame, $v$ is a constant velocity in pixels per frame (PPF), $m$ is the eye-integration index for smooth tracking.

We make the following assumptions: 1) the observer tracks the object perfectly with smooth movement of the eye balls; 2) the light stimulus within each frame is perfectly integrated in the human visual system; 3) each pixel of the display has identical temporal response behavior, i.e.,

$Y(x_S + m, t)$ is equal to $Y_0(t)$. Thus, Eq. (2) can be transformed to Eq. (3):

$$V(x_R) = \frac{1}{T} \int_{-\frac{x_R}{v}T}^{-\frac{x_R}{v}T+T} Y_0(t')dt' \tag{3}$$

From Eq. (3), the luminance profile of the reproduced blurred image can be expressed as a function of time. Let the sampling rate of the light measurement detector (LMD) be $R$. Then the total number of samples in one frame is $N = RT$. So, Eq. (3) can be transformed into Eq. (4), which is commonly used:

$$V(k) = \frac{1}{N} \sum_{i=k}^{k+N} Y_0(t) \tag{4}$$

$V(k)$ is derived from the liquid crystal response curve (LCRC) by applying one-frame-time moving-window function convolution [6]. $V(k)$ corresponds to the moving picture response curve (MPRC), which can be used to derive the BEW, N-BET, and MPRT described in Eq. (1). Figure 3 illustrates the relationship between the LCRC and the MPRC obtained using a one-frame-time width convolution.

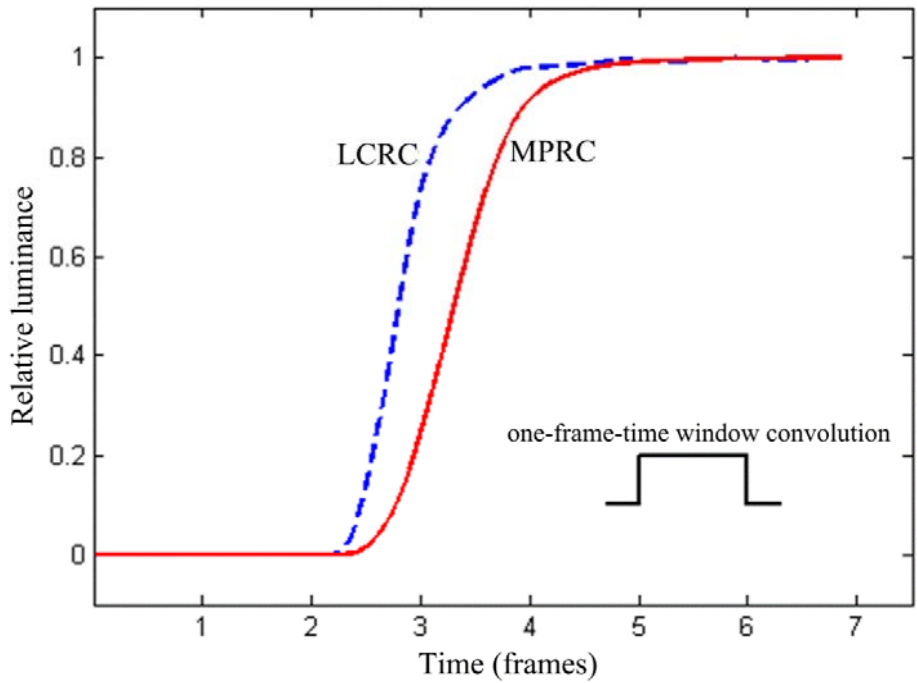

Figure 3- The relationship between LCRC and MPRC obtained using a one-frame-time moving window function convolution.

## 2.4 Motion blur from moving block width

As shown in Figure 1, different displays lead to different perceived width in retina coordinates when displaying dynamic images. Thus, the perceived width can be used to characterize the motion blur (similar to using BEW).

BEW measures the width of the blurred edge from 10 to 90% of the luminance transition, whereas the perceived width we proposed is the width of the whole moving block. For example, a static pattern with a width of 500 pixels moves from left to right on the screen leading to a perceived width different from 500 pixels when in motion.

A relative luminance of 90% is an imperceptible threshold in the case of a dark moving object and a relative luminance of 10% is an imperceptible threshold for a bright moving object. As shown in Figure 2, MBW is the width of $\overline{ad}$ and indicates the number of pixels exceeding the 10% luminance threshold on the luminance curve (see Figure 2). Though these metric definitions are similar, they are obtained from different sources. BEW is derived from the frame convolution of the LCRC and MBW is the distance that the moving block covers over the area measured by the LMD.

A greater object velocity leads to a more significant motion blur. However, according to Eq. (4) and BEW, motion blur is quantified by a time-relative metric rather than the object velocity. Thus, these metrics are not sufficient for comprehensive motion blur analysis. Our proposed MBW metric can assess motion blur at different velocities in various types of displays.

# 3 Experimental setup

In this section we describe our set-up to record the MBW. A schematic representation of the measurement setup is shown in Figure 4.

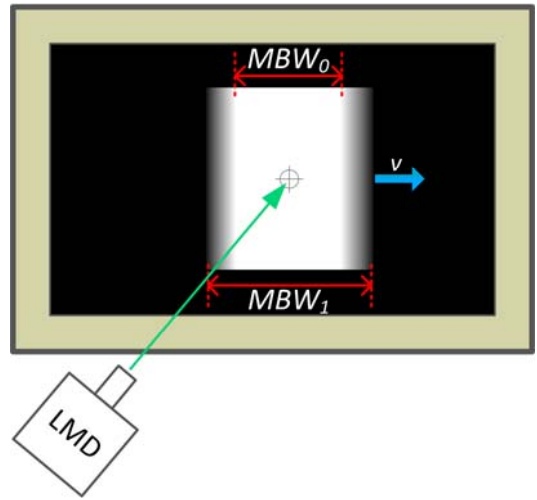

Figure 4- Measurement setup for MBW. The red line indicates the width of the test pattern. $MBW_0$ is the width of the block at stationary state and $MBW_1$ is the width when moving.

Our LMD is a PR-680L supplied with measuring apertures (1°) in the viewing distance 3 times of the display height. The detected area of the LMD (green arrow pointed) contains just more than 500 pixels, which is recommended by IDMS [5]. And an automated measurement shutter with a sampling rate 20 kHz is equipped.

The input signal to the display under test (DUT) should provide a stable level to ensure that the input is as designed. We require a programmable signal generator to enable object velocity changes as needed. The programmable video signal generator VG-870B meets these requirements since it can generate a stable video signal, which fits the VESA signal specification standard and it allows for arbitrary test patterns. The image resolution inputted to the DUT is set to $1024 \times 768$, and the frame rate is set to 60 Hz, and the static MBW is 512 pixels.

The block moving process is shown in Figure 4. When the white block moving from left to right periodically, it will cross the screen center, and then we can obtain the time *t* elapsed for the white block to traverse the display center (green arrow) by recording the luminance of the central pixels, i.e., for one period, the luminance of the central pixels will rise up from black to white, then sustained until the block's left edge cross the center, then the luminance will fall down. We then derive the pixel MBW of the white block and the change in width caused by

motion ($\Delta MBW$):

$$\begin{aligned} \Delta MBW &= MBW_1 - MBW_0 \\ &= t \times v - w_0 \end{aligned} \quad (5)$$

where $v$ is the scrolling velocity in PPF, $t$ is the crossing time in seconds divided by the frame rate (60 Hz) to obtain frames, and $MBW_1$ is the moving block width, $MBW_0$ ($w_0$) is the static block width.

Finally, $\Delta MBW$ would become a metric for the proposed method. For a specific speed, we could utilize $\Delta MBW$ to make a comparison between different display types. Moreover, only one specific speed may not be sufficient for characterizing the display motion performance. Therefore, we would like to merge different $\Delta MBW$ corresponding to different speeds into an overall metric, i.e., the slope of the $\Delta MBW$-speed plot.

The proposed method is accurate yet also very simple and fast to implement, resulting in a convenient alternative to the existing methods mentioned above. To prove the accuracy and feasibility of this method, we provide MBW experimental results from simulations and perception experiments.

# 4 Perception experiment

The purpose of our perception experiment is to compare the dynamic display performance of various types of displays at different scroll velocities. This direct comparison allows us to rank the motion performance of the DUTs.

## 4.1 Design and procedure

Participants took part in two separate testing sessions for two different reference scroll speeds. We had the same eight participants (five women and three men, with an average age of 22 years) through both sessions. The participants were screened for visual acuity and color vision using the Snellen chart and the Ishihara test, respectively. As seen in Figure 5(b), the ambient illumination is similar to typical room lighting [14], and the participants choose whether they sit or stand.

We test nine commercial DUTs in this experiment; the specifications are shown in Table 1. LCD02 and LCD05 are the same model so that we may judge any differences present between the same types of LCDs. The participants were asked to compare the reference block (top) with the velocity-changing block (bottom) as shown in Figure 5(a). A black screen is shown for 5 seconds between each successive stimulus to allow the eyes to adapt.

Motion blur is imperceptible with slow scroll speeds and extremely fast scroll speeds are not common in the real world. Thus, we set the scroll speed to a value between 5 PPF and 20 PPF. The velocity-changing block rendered at the bottom of the DUT has a random velocity and the reference block at the top of the DUT has a constant velocity. The size of the static test pattern is $256 \times 288$ pixels.

Table 1- DUT specifications

| No. | ID | Specifications |
|---|---|---|
| 1 | CRT01 | 1600 x 1200, 19" |
| 2 | PDP01 | 1920 x 1080, 60" |
| 3 | PDP02 | 3D display, 1024 x 768, 43" |
| 4 | LCD01 | 3D display, LED without backlight blinking, 40" |
| 5 | LCD02 | TN, LED with backlight blinking, 22" |
| 6 | LCD03 | IPS, LED without backlight blinking, 23" |
| 7 | LCD04 | TN, LED with light bar system, 19" |
| 8 | LCD05 | TN, LED with backlight blinking, 22" |
| 9 | LCD06 | TN, CCFL without backlight blinking, 19" |

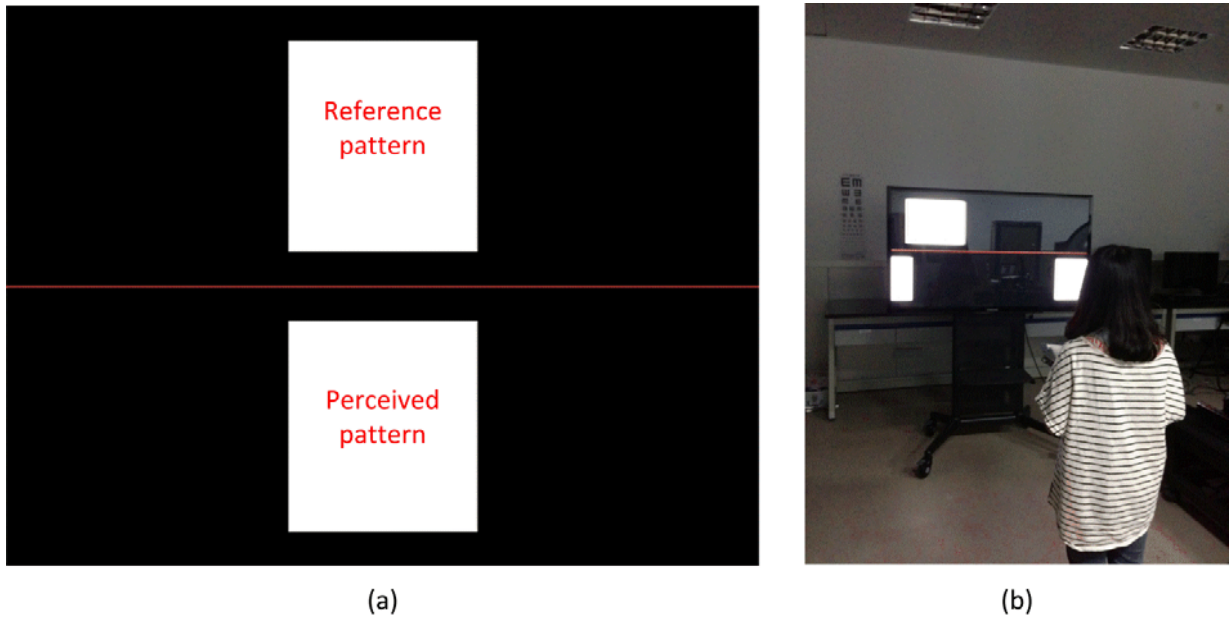

Figure 5- (a) A schematic of the paired comparison perception experiment. The top block is the reference with a constant scroll speed and the bottom one assumes a random scroll speed between 5 and 20 PPF; (b) The testing environment.

## 4.2 Session 1: 12 PPF reference speed

For measuring perceived motion block blur, a questionnaire method is used. The participants are asked to give scores on the perceived pattern comparing with the reference pattern.

In this session, the reference block speed was set to 12 PPF (the median of 5 and 20 PPF). The purpose is to compare the motion blur of the bottom block to the top one using the 7-point Likert scale. A positive score means the moving edge of the bottom block is clearer than that of the top block (i.e., a higher score indicates a better clarity of the perceived edge). A negative score indicates that the bottom block's edge is more blurred than the top block's edge.

If the blur of the perceived pattern showing in the bottom is viewed to be more serious than that of the reference pattern showing on the upper part, the participant would like to mark a negative score according to Table 2, vice versa for the perceived pattern less serious than the reference pattern, and then get a positive score.

Table 2- Blur evaluation scale for session 1

| Description | Much worse | Worse | Slightly worse | The same | Slightly better | Better | Much better |
|---|---|---|---|---|---|---|---|
| Score | -3 | -2 | -1 | 0 | 1 | 2 | 3 |

We use the participants' evaluations to score the DUTs. The lower the DUT's score, the more persistent the motion blur. Following this first session, the participants would have a whole understanding on the degree of motion blur for 9 DUTs at different velocities, i.e., became familiarized with all the level of motion blur for all cases. Then in the next perception experiment with reference pattern in 0 ppf, the participants can put motion blur from 9 DUTs at different velocities into different grades. A higher level, i.e., higher score means better dynamic display performance.

### 4.3 Session 2: Static reference pattern

In this session, the reference block was static, and the bottom block was set to a random velocity between 5 and 20 PPF. We change the score scale to account for the static reference block as shown in Table 3. A lower DUT score indicates a poorer dynamic performance.

Table 3- Blur evaluation scale for session 2

| Description | Extreme blur | Very blur | Much blur | Rather blur | Blur | Slightly blur | The same |
|---|---|---|---|---|---|---|---|
| Score | -6 | -5 | -4 | -3 | -2 | -1 | 0 |

The results of this perception experiment are shown in Figure 6 (the higher the score, the better the motion performance, i.e., less serious motion blur). The CRT has the best dynamic performance, but the motion blur from the PDPs and LCDs is difficult to distinguish as human perception only allows for an overall evaluation of whether the display is fast or slow [15]. To see a more precise distinction, we propose the simulation and camera methods. In the next section, we show a comparison of the motion blur using MBW, BET, and the human perception questionnaire score.

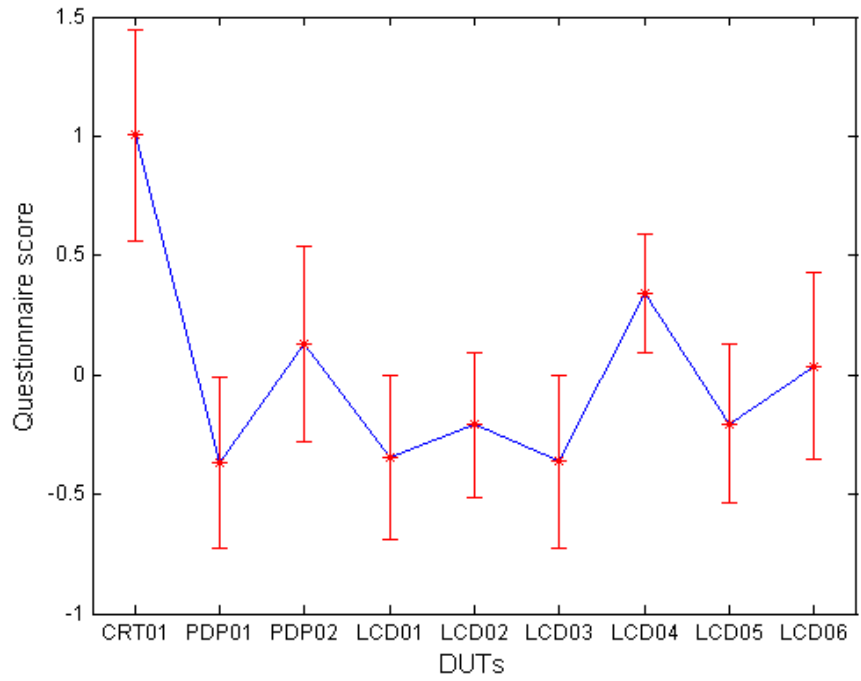

Figure 6- The perceived motion blur for CRT, PDP, and LCD. The higher the score, the less blurred the edge. The red error bar is the standard deviation.

# 5 Results

## 5.1 Motion blur from MBW

Using our method, the width of the entire moving pattern is used to quantify the motion blur. Differences in observed MBWs are attributed to varying scroll speeds or the same scroll speed viewed on different DUTs. Figure 7 shows the MBW for the nine DUTs for the various scroll velocities. To prepare the visual comparison of the different DUTs, we use a linear regression of all the velocities:

$$y = a + bx \qquad (6)$$

where, y is the MBW, $b$ is the slope, $x$ is the velocity, and $a$ is the y-intercept. Figure 7(a) illustrates the regression result for LCD03 for a square test pattern with 512 pixels, 0-255 grayscale, and movement to the right (these conditions are different from those used for LCD03 in Figure 7(b)). Figure 7(b) shows the comparison of all nine DUTs under the same conditions.

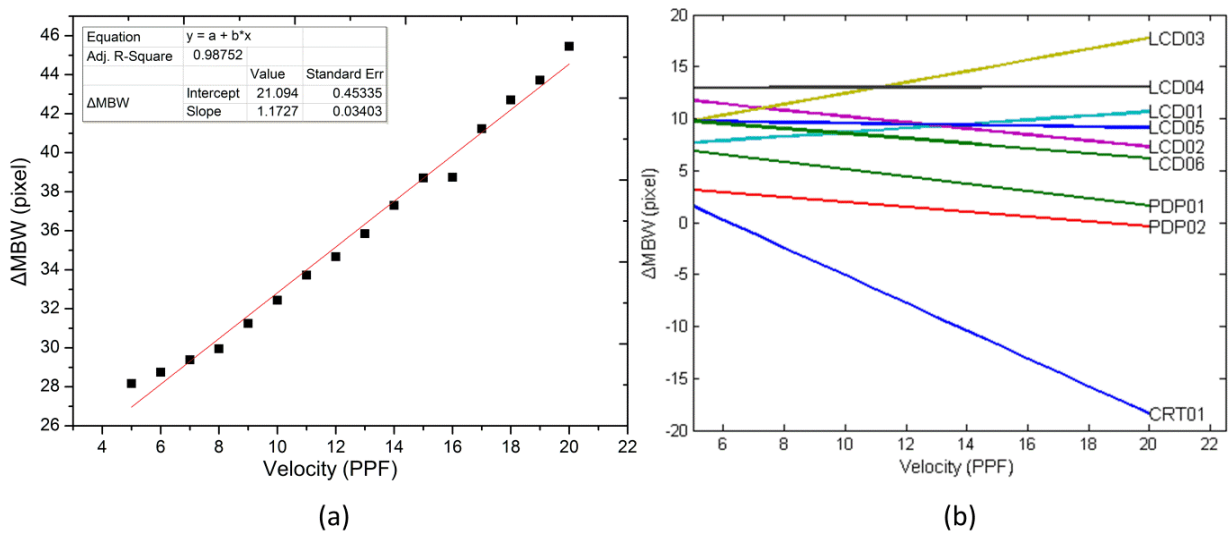

(a)            (b)

Figure 7- MBW as a function of scrolling velocity. (a) Regression analysis for LCD03; (b) Comparison of motion blur for nine DUTs.

In Figure 7(b), a lower value of b (the slope) indicates better dynamic performance. CRT01 has by far the most negative slope indicating it has the minimal motion blur of the nine DUTs. Its linear regression function is thus a paragon for the other displays to approach better motion performance.

It is not sufficient to rely only on the slope to characterize the motion blur, the intersection between the regression functions of two DUTs is also significant. Motion blur can be more severe in one display than another for low velocities but less severe than another display at high velocities. Motion blur does not necessarily exhibit a linear correlation to scroll velocity. Our MBW method allows us to provide a more detailed analysis of motion blur given different scroll speeds.

## 5.2 Comparison of motion blur

To validate our MBW method, we compare the motion blur according to MBW with the motion blur from the simulation approach and the perception experiment. Though the camera method is also a suitable reference for comparison, we use only the simulation method since the MPRT values obtained by these two methods are highly correlated [2], [16].

Figure 8 shows the BET calculation results from the simulation approach. We show two types of MPRCs: the first is the frame width convolution of the original LCRC (without numerical treatment); the second is the convolution of the improved LCRC (convolution corresponding to the backlight modulation frequency [17]).

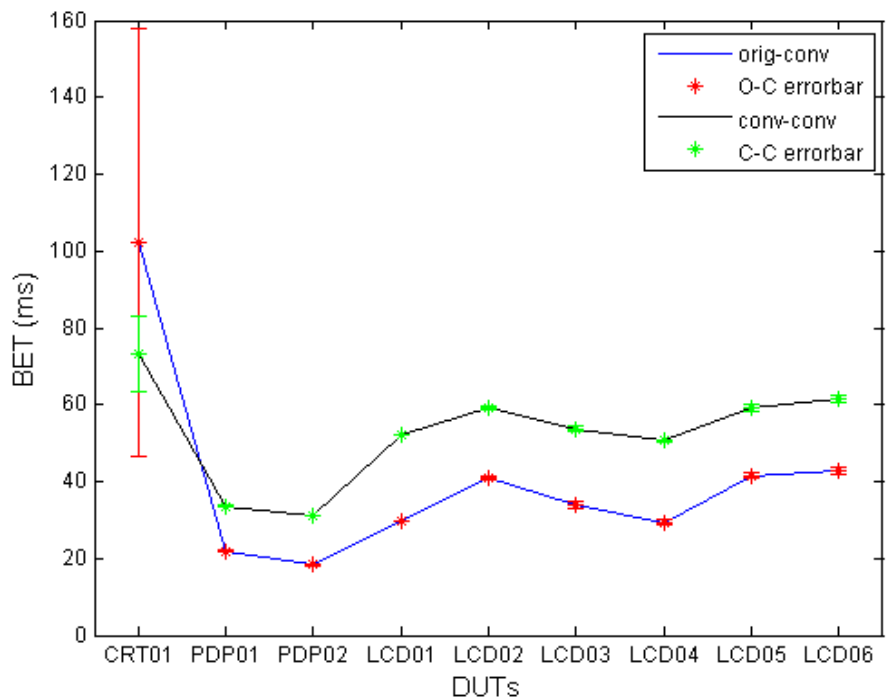

Figure 8- Motion blur obtained through simulation. The blue and black lines correspond to the original LCRC (orig-conv) and the improved LCRC (conv-conv) one frame convolutions, respectively. The red and green error bars show standard deviation.

It is clear that the BET from the improved LCRC (black line) is more stable, especially for CRT01. Thus, we use it in the final comparison of the three motion blur evaluation methods.

Since the BET, questionnaire score, and MBW have different units, we transform the three metrics into standardized z-scores. The z-score comparisons are shown in Figure 9. We thus obtain the rank of the dynamic display performance with each method.

$$z-score = (x - mean(x))/std(x) \tag{7}$$

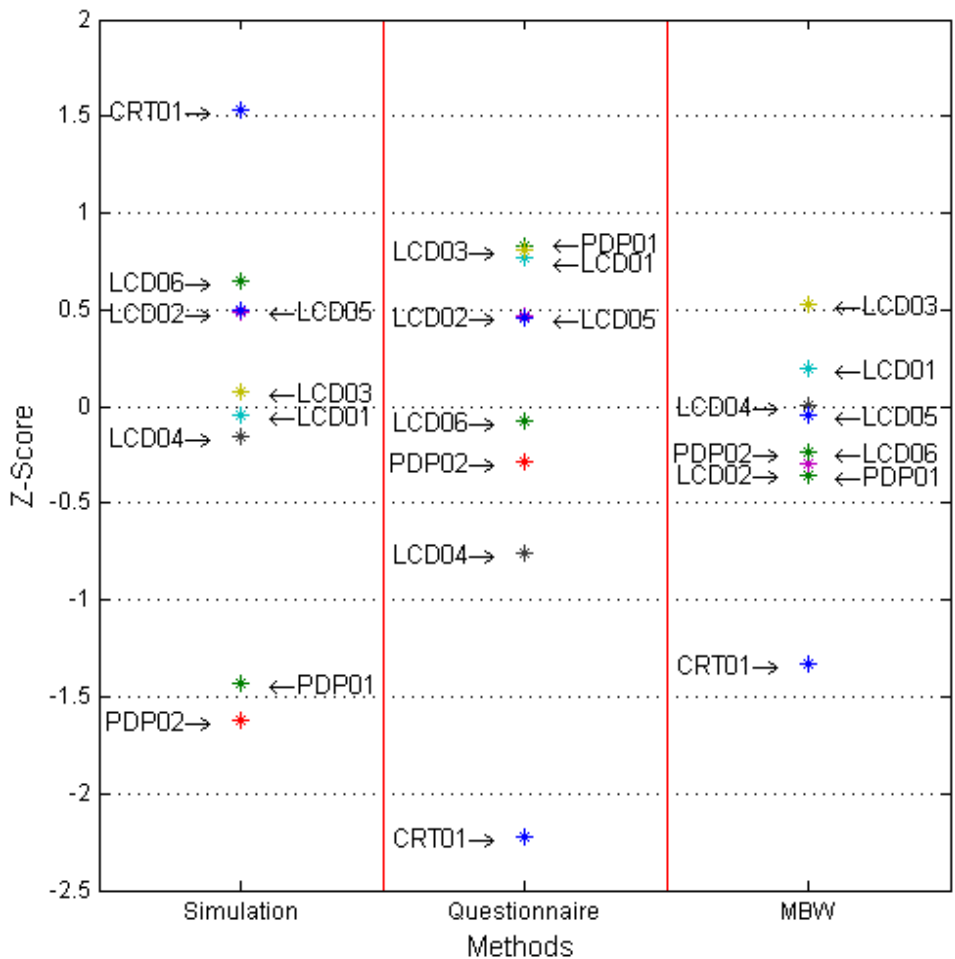

Figure 9- Standardized z-score comparison of BET, questionnaire score, and MBW.

Figure 9 shows the motion blur obtained from the three methods. We summarize the results as follows:

- The simulation algorithm is only appropriate for characterizing LCD motion blur, when it is applied to PDPs and CRTs, the simulation model needs improvement [18], [19].

- While we cannot guarantee the precision of human judgment in the questionnaire, it does aid in validating the MBW since the z-scores are similar. Our MBW method is applicable to more situations than the simulation method because it can be used to quantify the motion blur not only for LCDs, but also for PDPs and CRTs.
- The DUT motion blur results can be divided into groups where the motion performance is similar within each group.

# 6 Discussion

Although MBW and BEW both signify the width of the object, the two concepts differ. BEW is the width of the blurred edge alone, which is derived from the 10–90% contrast change using a response curve. MBW is the width of the entire moving block and is a metric related to the scrolling velocity and the display response of the measured pixels.

A small change in MBW is preferred as the speed changes. The MBW for an ideal display would be constant. CRT01 showed the best dynamic performance and the MBW decreased as the scroll speed increased. This makes sense as a result of scanning and speed increments. A higher velocity indicates a greater span of gray level variation (the frame-by-frame position is $x \to x+v$). Since the measured area includes more than 500 pixels in the center of the screen, and the span of motion can be significant, the LMD may overlook part of the luminance of the moving block in a frame. This situation generally occurs when the moving block is entering or exiting the test area (at the extremities of the display test area). Moreover about the detected area, two points need further discussion. One is the edge pixel luminance when the edge of the moving block is overlapping the detected area. Another is the size of the detected area.

For the edge pixel luminance effect, to totally eliminate the effect of edge pixel luminance, only one pixel should be used as the detected area theoretically. However, only one pixel is impossible for LMD to record, so that a detected area contains more than 500 pixels is used to make an approximation to this theoretical one pixel. The results shows by adopting the 500 pixel detect area, the effect of edge pixel luminance is minor. Thus we can just consider that this is only one discrete point.

The luminance of the detected area may be affected by the size of the detected area. However, only the relative value of the luminance (10% or 90%) is used to determine the luminance threshold for MBW evaluation, rather than the absolute value of the luminance. Therefore, in this aspect, the variation of the detected area has no effect on the measured results. And of course on the premise of that the size of the moving block should be far larger than that of the detected area.

To obtain an accurate BET, the LCRC must be improved by using moving-average-window convolution filters to filter out the backlight modulation and random noise. However, this is not crucial when using our MBW method. A comparison of the results before and after of such a convolution using the MBW method shows only slight differences and the motion blur rank of the nine DUTs remains unchanged. This is attributed to the difference between the measurement method for LCRC and MBW. Though they are measured with the same LMD, the test pattern for LCRC is composed of periodic changes whereas a moving block serves as the test pattern for the MBW method. Thus, the MBW method is more robust even without convolution.

Moreover, in current study we focus on describing the MBW method for different display types and different moving speeds. The gray scale used in the experiment is only black-to-white. Of course, gray scale is another variable and gray-to-gray motion blur characterization may be more interesting. However, just like the gray-to-gray response time, there is no doubt that the proposed method can obtain the luminance profile from different gray scales. The MBW method is still applicable for any other gray levels.

To improve the feasibility and applicability of the MBW method, future work should focus on determining the relationship between motion blur and the inputted signal (e.g., the effect of the resolution, refresh rate, size, gray-scale, shape, and direction of the moving pattern). Some preliminary results show that motion blur is not only influenced by velocity, but also by the gray levels of the test pattern. The initial $\Delta MBW$ for a 5 PPF velocity is used as a reference to evaluate motion blur and the slope of the regression fit line for increasing velocity shows the effect on motion blur for each display.

# 7 Conclusion

Motion blur is a key dynamic performance measure of image quality for different displays. However, existing motion blur measurement methods were designed for LCDs. In our work, we propose a straightforward method to quickly and simply quantify motion blur of any type of display, particularly, CRT, PDP, and LCD. Furthermore, our method analyzes the velocity characteristic that are neglected when using BET. Perceptual experiments using a moving block with a changing velocity validate our evaluation method. The results indicate that the proposed MBW method can quantify motion blur for various types of displays in a robust and comprehensive way.